\documentclass[runningheads]{llncs}
\usepackage[T1]{fontenc}
\usepackage{tabularx}
\usepackage{graphicx}
\usepackage{booktabs}
\usepackage{caption}
\usepackage{subcaption}
\usepackage{hyperref}
\begin{document}
\title{Extending 6D Object Pose Estimators for Stereo Vision}
%
%
\author{Thomas Pöllabauer\inst{\star,1,2}\orcidID{0000-0003-0075-1181} \and
Jan Emrich\inst{\star,1,2}\orcidID{0009-0008-2883-0555}  \and
Volker Knauthe\inst{2}\orcidID{0000-0001-6993-5099}\and
Arjan Kuijper\inst{1,2}\orcidID{0000-0002-6413-0061} \newline
$\star$ contributed equally}
\authorrunning{Thomas Pöllabauer et al.}
%
\institute{Fraunhofer Institute for Computer Graphics Research IGD, Darmstadt, Fraunhoferstraße 5, Darmstadt, Germany \and
Technical University Darmstadt, Karolinenplatz 5, Darmstadt, Germany
\email{thomas.poellabauer@igd.fraunhofer.de \\
mail@janemrich.de}}
\maketitle              
\begin{abstract}
Estimating the 6D pose of objects accurately, quickly, and robustly remains a difficult task. However, recent methods for directly regressing poses from RGB images using dense features have achieved state-of-the-art results. Stereo vision, which provides an additional perspective on the object, can help reduce pose ambiguity and occlusion. Moreover, stereo can directly infer the distance of an object, while mono-vision requires internalized knowledge of the object's size. To extend the state-of-the-art in 6D object pose estimation to stereo, we created a BOP compatible stereo version of the YCB-V dataset called YCB-V DS and extended multiple state-of-the-art pose estimator algorithms to be applicable to stereo. Our method outperforms state-of-the-art 6D pose estimation algorithms by utilizing stereo vision and can easily be adopted for other dense feature-based algorithms.

\end{abstract}

\section{Introduction}
The task of 6D object pose estimation (6DOPE) involves determining the translation and rotation of an object within a scene. This is a crucial task in computer vision with applications in various fields, including robotics, quality control, and augmented reality. In robotics and quality control, precise object pose estimation is necessary to grasp, move, or apply manufacturing processes to objects, or to ensure their proper position during assembly. In augmented reality, accurate object pose estimation is required to merge the real world with overlaid information or virtual worlds. In all these fields, imprecise or inaccurate pose estimates, as well as low frame rates, can render tasks impossible or lead to a poor user experience. \\
To achieve high accuracy in real-time applications, such as robotics, traditional methods often rely on specialized depth cameras using time-of-flight or projected light sensors to quickly capture images. While these sensors offer fast performance, they can also introduce noise and are prone to failure when dealing with reflective or transparent objects. Also, direct regression methods recently attained the highest level of accuracy for 6DOPE. These methods rely on powerful dense intermediate representations, such as 2D-3D correspondences, which enable the prediction of the 3D coordinates related to each pixel in the image plane. With these correspondences, pose estimation can be iteratively calculated using algorithms like RANSAC. In recent developments, this iterative process has been replaced by neural networks, facilitating end-to-end training and faster inference. By directly estimating the pose in a single step, pose estimators can meet real-time requirements without the need for iterative algorithms. \\
Finally, in an effort to increase accuracy of pose estimators, previous work utilized multiple views. Stereo is an intuitive special case of multi-view inspired by human vision offering an additional perspective and indirect depth images, which can help overcome the size ambiguity issue in monocular object pose estimation. Previous work on keypoint methods have demonstrated improved accuracy adopting stereo when compared to their monocular versions highlighting the potential of stereo imaging for 6DOPE. \\

In this work, we test the assumption that extending a monocular pose estimator to stereo vision is an effective way to greater pose accuracy and we test this hypothesis using current state-of-the-art correspondence-based algorithms. In particular our contributions include:
\begin{itemize}
    \item an extension of the state-of-the-art pose estimator GDRNet for stereo cameras
    \item YCB-V DS, a stereo version of the relevant YCB-V object dataset including depth information
\end{itemize}

This work is structured as follows: First, we will discuss related work. Then, we will present our dense stereo 6DOPE method and reasoning for its design, motivating the applicability to other, similar algorithms. Next we introduce our stereo YCB-V dataset (YCB-V DS) used to train and evaluate our method. Finally, we evaluate all proposed approaches and compare our method to state-of-the-art end-to-end 6DOPE. 

\section{Related Work}
In the field of 6DOPE, depth information was traditionally heavily relied upon for achieving the highest accuracy through refinement. However, with the widespread availability of conventional cameras, deep learning on RGB images has gained significant interest. In the following, we will describe the common methods used for 6DOPE.

\subsection{Keypoint Methods}
Keypoint methods involve detecting a set of characteristic points on the surface of the target object. The pose is then estimated by solving a Perspective-N-Point (P$n$P) problem, typically using the RANSAC algorithm. Historically, keypoints were detected using the SIFT \cite{lowe2004distinctive}, SURF \cite{surf} or similar handcrafted features, while today keypoints are predicted using neural networks. One recent method doing so is HybridPose \cite{song2020hybridpose}. Keypoint methods can easily be extended to the stereo modality when framing the P$n$P problem. Also, keypoint-based methods have already been adapted for stereo pose estimation, with accuracy gains ranging from 14 to 25 percentage points over their monocular counterparts. \cite{liu2020keypose,peng2019pvnet,liu2021stereobj}

\subsection{Iterative Pose Refinement}
We differentiate two subcategories for iterative refinement methods: Refinement in image space, and refinement in point cloud space.
Iterative Closest Point (ICP) is a widely used method and falls into the latter category requiring a point cloud. It enhances the object's pose estimation by matching the predicted point cloud to the available geometry information. The method is often applied to depth data after an initial pose estimation with another method due to its ability to achieve high accuracy given sufficient inference time and good depth data.
In image space-based refinement methods, the initial pose estimation is refined iteratively by rendering the best current guess and predicting the difference between the actual pose and the estimated pose in the input image. This process is repeated a set number of times. DeepIM \cite{li2018deepim} was the first method to introduce such refinement, while CosyPose \cite{labbe2020cosypose} combined the DeepIM approach with scene-level refinement. Coupled Iterative Refinement \cite{lipson2022coupled} refines both pose and correspondences in lockstep. RePose \cite{repose} achieves faster run-time by replacing the repeated forward passes of a CNN-based optimization with a fast renderer and a learned 3D texture.

\subsection{End-to-end dense Feature 6DOPE}
\textbf{GDR-Net.} To the best of our knowledge GDR-Net (GDRN) \cite{wang2021gdr} was the first end-to-end trainable network that directly regresses pose using intermediate dense features. Unlike previous methodologies, GDRN harnesses dense representations to enhance predictions without the need for a second stage, such as RANSAC, to obtain final results. GDRN uses a 3 stage pipeline:
First, an object detector is employed to identify the bounding box and class of the target object. Second, the bounding box-cropped image undergoes processing through the geometric feature regression network, utilizing an encoder-decoder architecture. In the case of GDRN this is based on ResNet \cite{he2016deep}, while a modified version named GDRNPP uses ConvNeXt \cite{convnext}. The decoder predicts various features, including the object mask, 2D-3D correspondences, and surface regions. The 2D-3D correspondences are regressed into three layers, each representing a dimension of the object coordinates.
Third, the concatenated 2D-3D correspondences and surface regions are fed into the Patch-PnP network, composed of convolutional layers and two dense heads for regressing the translation and rotation of the pose. 

\textbf{SO-Pose}. SO-Pose \cite{di2021so} extends the capabilities of GDRN by introducing an additional intermediate dense feature, self-occlusion. In the neural network architecture, a new decoder head is incorporated into the second stage of the network. This added component predicts occlusion correspondences, providing information about the self-occlusion of the 3D object. Specifically, for each origin plane of the object coordinate system, the corresponding other two dimensions are predicted, resulting in six layers. The self-occlusion maps are designed to offer supplementary insights into the 3D object. This information, being independent of the noise present in the 2D-3D correspondences, should help with ambiguous poses and textureless objects. 

\subsection{Disparity and Stereo}
Stereo Matching, a well-established task in Computer Vision, involves estimating the disparity map in horizontally aligned stereo images. Disparity represents the distance between corresponding pixels in two stereo images, indicating the depth in the 3D world. 

Zbontar and LeCun \cite{zbontar2015computing}, along with Zagoruyko and Komodakis \cite{zagoruyko2015learning}, pioneered the use of Convolutional Neural Networks (CNNs) for stereo matching, employing a siamese network that concatenated feature representations from two images to predict a disparity map.
Luo, Schwing, and Urtasun \cite{luo2016efficient} advanced this model by computing the inner product of image features instead of concatenation, leading to improved matching accuracy and accelerated inference times.
Shamsafar et al. \cite{shamsafar2022mobilestereonet} built upon these works by incorporating 2D MobileNet \cite{sandler2018mobilenetv2} blocks into stereo matching, reducing computational costs while maintaining high accuracy. Additionally, they introduced an innovative "Interlacing Cost Volume Construction" technique.

Utilizing disparity maps for obtaining depth from stereo images also has evolved with the success of deep-learning methods. The emergence of end-to-end networks, exemplified by references \cite{mayer2016large,kendall2017end,du2019amnet,yang2018segstereo}, has been influential in predicting disparity maps. These networks typically incorporate correlation or distance cost volumes, a concept initially introduced by FlowNet \cite{fischer2015flownet}.
KeyPose \cite{liu2020keypose} introduced an early fusion architecture for stereo pose estimation, leveraging a disparity map to directly predict the depth of keypoints. Furthermore, the dense PV-Net \cite{peng2019pvnet} can be applied to stereo scenarios by predicting keypoints for each image and subsequently triangulating the object using keypoints from both images.

\subsection{BOP Challenge}
The evaluation of methods for rigid object pose estimation faces challenges due to variations in datasets, including differences in lighting, occlusion, object textures, and spatial distributions of test poses. To address these issues, \cite{hodan2018bop} introduced a comprehensive benchmark for Mono RGB-D images using relevant metrics for 6DOPE \cite{hodavn2016evaluation}. Their contributions include selecting and transforming eight datasets into a standardized format, defining evaluation metrics, and introducing an evaluation system. Noteworthy datasets within the benchmark include the Linemod and Linemod-Occluded datasets and the YCB-Video dataset. Additionally, they provided results and discussions on methods across all datasets. To facilitate research on this benchmark, they also released the BOP Toolkit \cite{bop_toolkit}. In subsequent years re-occurring challenges continued to survey the state of the art \cite{hodavn2020bop,sundermeyer2023bop,bop23}.

\section{Approach}

In the field of end-to-end 6DOPE, current state-of-the-art techniques are primarily focused on single images without an obvious methodology how to incorporate additional views. We propose a way to extend 6DOPE algorithms using dense features to the stereo modality by exploring various approaches to fuse information from two views and introducing disparity features. 

\begin{figure*}
    \centering
    \includegraphics[width=1\linewidth]{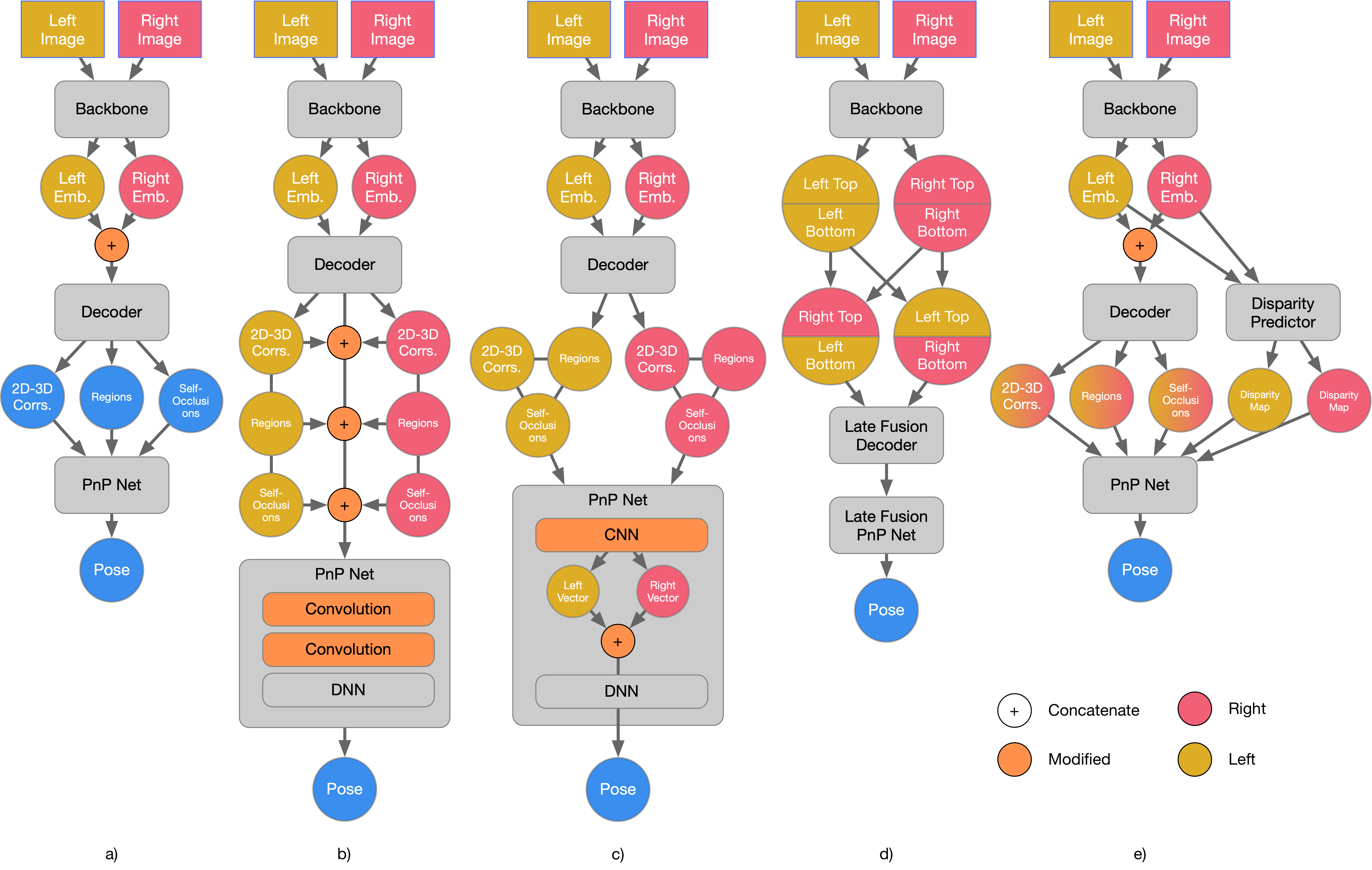}
    \caption{Details on some proposed architectures for feature fusion at different pipeline stages. Exemplified on the SO-Pose algorithm, but also applicable to GDRN. a) Early fusion. Feature maps are merged after individual forward propagation through the backbone. b) Mid-level fusion. After two separate forward passes the embeddings are concatenated and additional convolutional layers are added inside the $PnP$ Net to cope with the additional dimensionality. c) Late fusion. Features are fused only after the CNN inside the $PnP$ network but before the multi-layer perceptron / dense network. d) Double fusion. Here we re-use the late fusion $PnP$ Net and combine it with an embedding mixing scheme where half of the feature maps are swapped between left and right. This way each individual forward pass has information from both images. e) Early + shared backbone based disparity prediction. We add a disparity prediction using the feature maps as extracted from our common backbone. Disparity prediction is done both ways in a symmetric fashion. The additional features are again concatenated and fed to the $PnP$ network for pose regression.}
    \label{fig:big_picture}
\end{figure*}

In subsequent sections, we discuss different architectures for fusing images to estimate object pose. Fusion is the process of combining the signals of both images at different stages of the processing pipeline. More specifically, we evaluate the following fusion approaches: Early Fusion, Mid Fusion, Late Fusion, Double Fusion, as well as an extension with additional features. All of these architectures are depicted in Figure \ref{fig:big_picture} and will subsequently be discussed in detail.\\

Our method builds upon the processing pipelines as found in GDRN and SO-Pose and begins with the estimator receiving detections with bounding boxes. These bounding boxes define a region of interest (RoI) and zooming into the RoI we obtain our embeddings. The next stage in the network predicts dense features such as 2D-3D correspondences, regions, and occlusion labels. Finally, a regression network determines the translation and rotation of the object from these features. Our contribution consists in exploring additional features like disparity and integrating multiple views into the existing pipeline at various stages. \\

\textbf{Feature Fusion.}
One of the most significant challenges in 6DOPE is pose ambiguity, which we argue, can be mitigated by incorporating additional views. To address this, we propose and evaluate the influence of fusing the extracted features of two images at different stages of the pipeline. We continue by describing our reasoning for the different approaches, and their combination.

\subsection{Early Fusion}
We propose merging the embeddings from the first stage in such a way that the second stage can infer 2D-3D correspondences and region classification from multiple views. To this end the feature maps from the left and the right view get concatenated after the backbone but before the decoder as shown in Figure \ref{fig:big_picture} a). To maintain the same merged channel dimension, we halve the channel dimension of the left and right embeddings. This approach should allow the network to accurately identify the object in the image, particularly for objects that may appear similar from certain views but not from others. As a result, we assume improvements in region classification and 2D-3D correspondence estimation.

\subsection{Mid-Level Fusion}
In our approach, we predict intermediate features, such as 2D-3D correspondences, region classifications, and masks, for each view separately. These features are then concatenated into a single feature tensor. This combined tensor is then fed into a perspective-n-point P$n$P network, which employs a CNN to predict translation and rotation. Mid-level fusion runs the backbone and the decoder heads per view before concatenating the resulting features (Figure \ref{fig:big_picture} b). We have to adjust the convolution layers in the P$n$P network to accommodate our changes. By adopting mid fusion, the prediction head should benefit from a more diverse set of 2D-3D correspondences and region mappings, and should thereby reduce pose ambiguity.

\subsection{Late Fusion}
In Late fusion (Figure \ref{fig:big_picture} c) we predict the intermediate features for each view separately and merge them relatively late, within the P$n$P network. More specifically, we run the features through the convolutional layers of the P$n$P network before concatenating them. The resulting combined feature vector is then passed through the final fully connected layers. By employing late fusion, we again hope to see a reduction in pose ambiguity.

\subsection{Double Fusion}
In our double fusion approach (Figure \ref{fig:big_picture} d), we integrate both early and late fusion techniques. After the first network stage, the embeddings of the views are concatenated after the backbone, this time though we mix the feature maps of both views before feeding them into the decoder. The decoder and P$n$P Net share the same design as found in late fusion, merging the features after the convolutional layer, but before the multi-layer perecptron. By combining the benefits of improved intermediate features and the inclusion of more views from both fusion techniques, double fusion should offer a more robust estimation of object pose.

\subsection{Early Fusion with shared Backbone Disparity Prediction}
Stereo vision can also be described by one image and the disparity to another. We argue that combining early fusion with a disparity prediction might improve overall accuracy. Early fusion can enhance the quality of dense features by merging embeddings, which should reduce pose ambiguity and rotation error. Disparity, on the other hand, should mainly enhance the translation prediction by inferring depth. We therefore integrate a disparity feature prediction to give the P$n$P network two additional signals, the disparity prediction per view (Figure \ref{fig:big_picture} e). The disparity features behave like the output of the decoder heads and are fed to the P$n$P network. Addition of disparity loss should compel to learn depth-relevant features, which we argue, should improve the pose estimation quality.

\section{Evaluation}
We extend state-of-the-art methods, which includes GDRN, GDRNPP, and SO-Pose, by implementing a stereo version of SO-Pose and GDRN, called SO-Stereo and GDRN-Stereo, respectively. For comparability with standard BOP datasets, we extend the BlenderProc \cite{blenderproc} Pipeline to generate all necessary data and created a stereo PBR YCB-Video dataset called YCB-V DS together with real stereo recordings.

\subsection{Synthetic and real Dataset}
\begin{figure}
    \begin{subfigure}[t]{0.5\linewidth}
        \centering
        \includegraphics[width=1\linewidth]{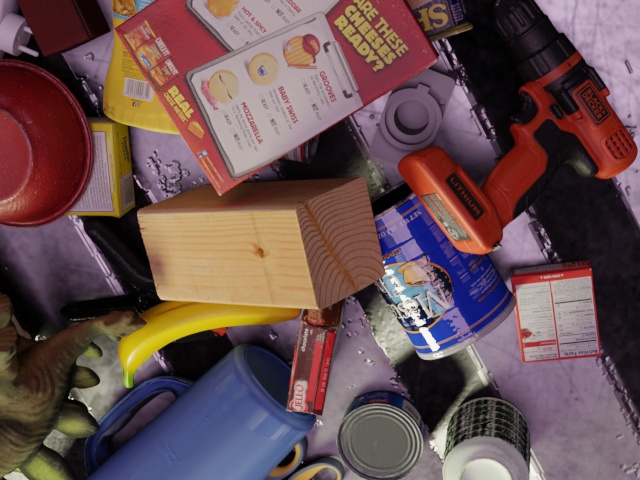}
        \caption{A sample taken from our PBR YCB-V DS dataset.}
        \label{fig:render}
    \end{subfigure}
    \begin{subfigure}[t]{0.5\linewidth}
        \centering
        \includegraphics[width=1\linewidth]{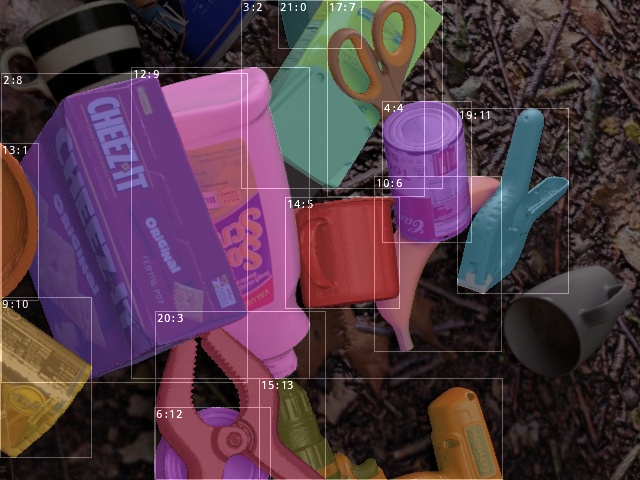}
        \caption{Another sample from our dataset. Perfect ground truth lends itself to compare performance between algorithms.}
        \label{fig:render_gt_bboxes}
    \end{subfigure}
    \caption{Physically-Based-Rendered Dataset Examples}
\label{fig:pbr_dataset}
\end{figure}
We created both stereo physically-based renderings in BOP format (Figure \ref{fig:pbr_dataset}), as well as a set of high quality real-world stereo + depth video recordings consisting of 32 scenes. \\

\textbf{Data Generation and Processing.} Our dataset includes a wide range of generated scenes with randomly sampled views, objects of interest, distraction objects, and object locations. We physically simulate the objects under gravity, which ensures a wide range of realistic occlusions in the resulting scenes. The background texture, lighting direction and intensity, viewpoints, material and reflectivity of objects were sampled randomly. By extending BlenderProc to output stereo pairs, we get perfect ground truth for the 6DOPE problem. 
In total, we rendered fifty thousand stereo frames consisting of RGB and depth images, including up to 15 target objects per frame. For each scene, 25 frames were obtained from random viewpoints of the scene. Annotations such as object masks and visibility percentages were computed with the BOP Toolkit. Next, the GDRN and SO-Pose repositories were used to compute additional required annotations, such as 2D-3D correspondences and occlusion correspondences. For computing 2D-3D correspondences we modified scripts from the SO-Pose repository, speeding them up using Numba, an LLVM-based Python JIT compiler, and pallelizing tasks with Python Multiprocessing.
In addition, we converted the file formats from the previous papers to a compressed file format (.npz) to reduce memory requirements and save space.
The SO-Pose occlusion labels were generated using scripts from the SO-Pose repository, and we again used Numba and Python Multiprocessing to accelerate the process and compress the files.
Finally, we removed ground truth labels if less than 10\% of the object surface was visible in either of the stereo image pairs from the dataset for stability reasons. We ended up with a training set containing 433.645 labels and a test set containing 48.080 labels. Also note that the average visibility of target objects in our dataset is only 62\%. High amount of occlusion is commonly recognized as increasing the difficulty of the 6DOPE problem, making our dataset quite challenging. \\

\begin{figure}
    \begin{subfigure}[t]{0.2025\linewidth}
        \centering
        \includegraphics[width=1\linewidth]{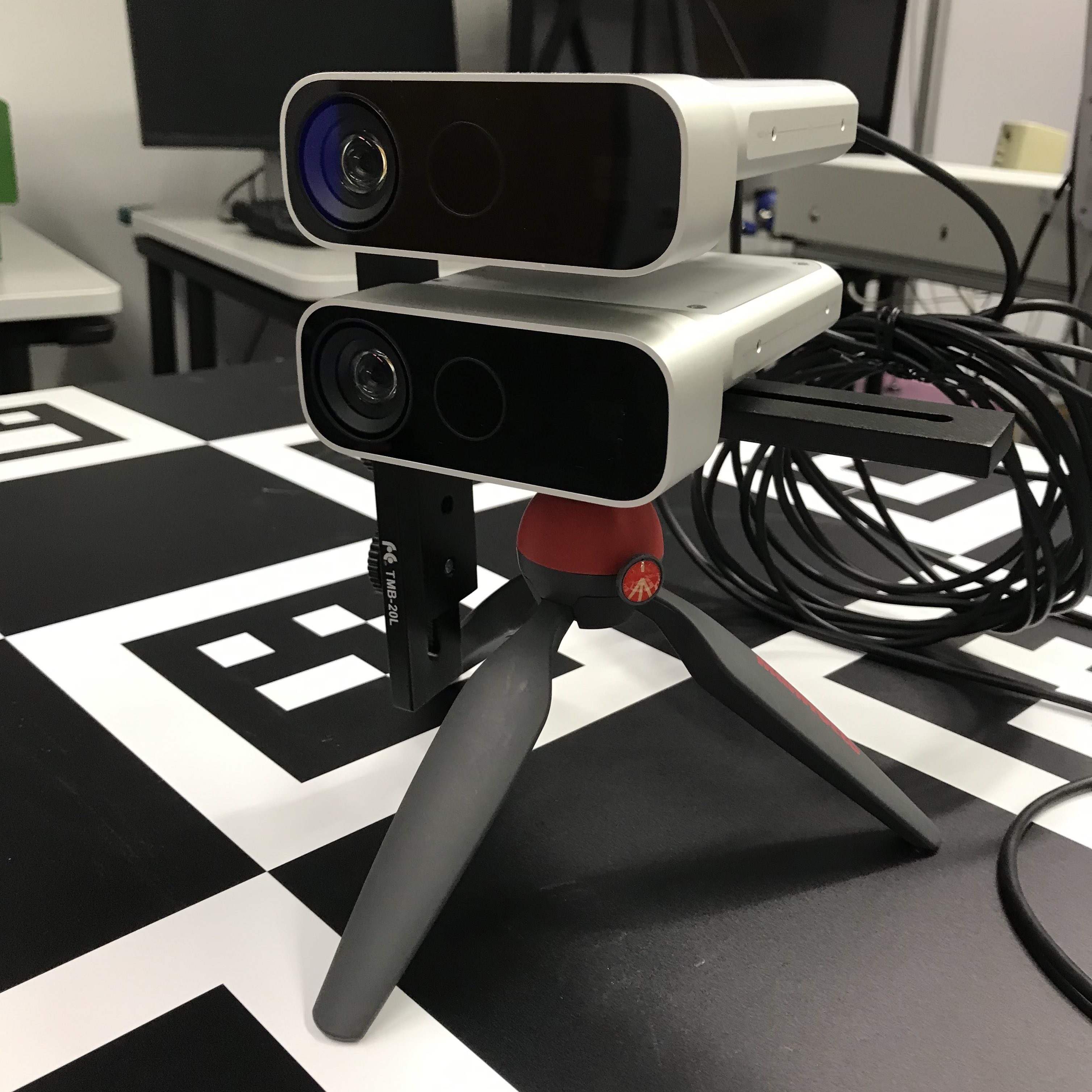}
    \end{subfigure}%
    \begin{subfigure}[t]{0.27\linewidth}
        \centering
        \includegraphics[width=1\linewidth]{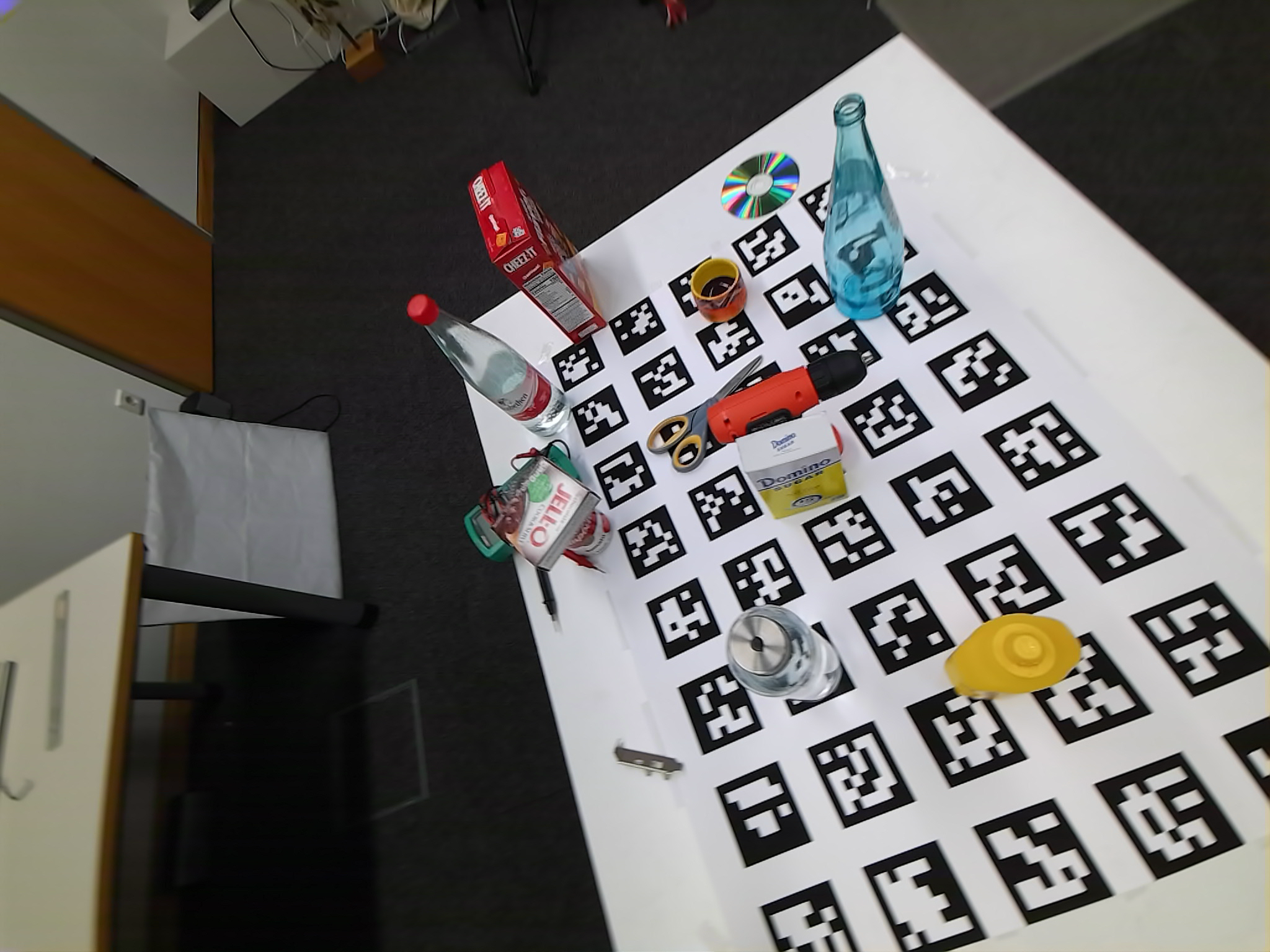}
    \end{subfigure}%
    \begin{subfigure}[t]{0.27\linewidth}
        \centering
        \includegraphics[width=1\linewidth]{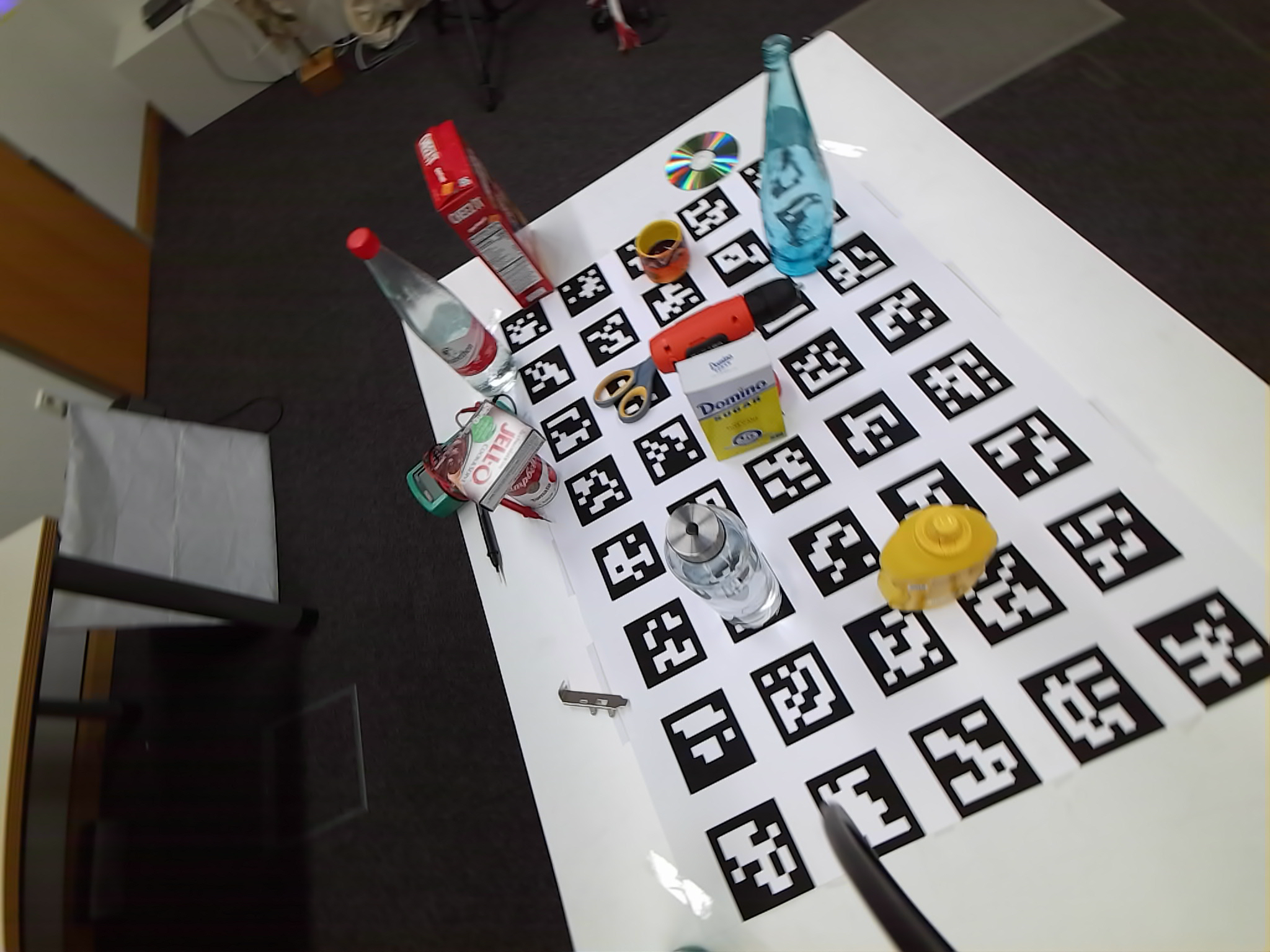}
    \end{subfigure}
    \begin{subfigure}[t]{0.225\linewidth}
        \centering
        \includegraphics[width=1\linewidth]{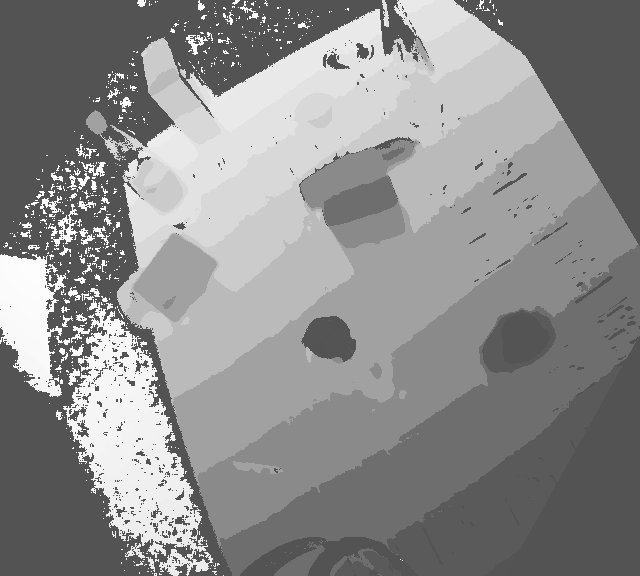}
    \end{subfigure}
    \caption{Using our setup of 2 Azure Kinect cameras we acquire high resolution color images together with depth information. From left to right: Our camera setup, frame of first camera, frame frame of second camera, one of the corresponding depth images.}
    \label{fig:real_data_sample}
\end{figure}

\textbf{Recordings.}
We provide videos (Figure \ref{fig:real_data_sample}) taken in three different sessions, including a wide range of object combinations and number of objects, as well as lighting conditions. Scenes include spacious and cluttered rooms, featuring large windows and diverse lighting conditions, including natural illumination, intense direct sunlight, artificial illumination, and a dimly lit room. For recording we used two Microsoft Azure Kinects mounted on a tabletop tripod with a baseline of 50 millimeters (Figure \ref{fig:real_data_sample}). Our setup exports two camera streams, each with a 2048 x 1536 pixel RGB signal plus a 640 x 576 pixel depth video.  \\
As for the test objects, unfortunately, some of the objects are not obtainable
anymore, thus they were replaced by newer versions. These objects differ
from the ones in the synthetic dataset. Specifically different are the objects 002\_master\_chef\_can (modified texture), 019\_pitcher\_base (color changed from blue to a transparent color with a red lid), and 035\_power\_drill (slight changes to overall appearance). Finally, we used various objects to distract object detectors and pose estimators. Among them are transparent or metallic objects such as glass water
bottles or pans.

\subsection{Parameterization}
Our approaches extend GDRN and SO-Pose, which we use as our monocular baseline. To make the comparison as fair as possible, we stick to their parameterization where possible. Since our data should be similar to the YCB-V dataset, we adopt the configuration specified by the authors, except for the adjustment of the parameter governing the exclusion of training parameters based on visibility: The visibility threshold, a training parameter initially set to 0.2 in the SO-Pose paper, dictates that training labels with less than 20\% visibility due to occlusion or cutoff are excluded. Since our synthetic dataset already excludes labels with less than 10\% visibility, and we aim to include cases where one label has 11\% visibility while the other has 22\%, we have removed this threshold across all configurations in the methods considered in our work.\\
Next we go into more detail on some of our considerations and parameters:

\textbf{Bounding Box Unification.}
The bounding boxes of the two images undergo expansion to encompass the bounding boxes from both images, resulting in a singular larger bounding box used for both images. Unification is essential for consolidating two detections into a single prediction, given that the prediction is made relative to the bounding box. Additionally, aligning the horizontal lines is necessary for stereo matching with depth estimation.

\textbf{Stereo Matching.} For obtaining general depth information, we utilize the disparity estimation network MobileStereoNet \cite{sandler2018mobilenetv2}. It is relatively lightweight, yet we additionally scaled the size of the hidden layers down to diminish memory requirements and reduce the computational cost of the network and shared the backbone with the rest of the network.

\textbf{Feature Heads.}
Except for the stereo matching features, the feature heads remain unchanged from the GDRN and SO-Pose baselines.

\textbf{Loss Function.}
The loss function encompasses the individual loss functions associated with GDRN, SO-Pose, and the stereo matching network (if used). We do not change the weighting of individual loss terms compared to the baseline algorithms.

\begin{table*}[]
\begin{tabular}{@{}llllllll@{}}
\toprule
Method                   & GDRN & GDRNPP   & SO-Pose & SO-Stereo     & \multicolumn{3}{l}{GDRN-Stereo}           \\ \cmidrule(l){6-8} 
Fusion                   &      &               &         & early         & \multicolumn{2}{l}{early} & double        \\ \cmidrule(lr){6-7}
Shared Backbone Disp. Pred.          &      &               &         &               & yes              & no     &               \\ \midrule
\textit{002\_master\_chef\_can}   & 48.7 & 48.7          & 20.1    & 29.0          & \textbf{53.0}    & 50.6   & 51.0          \\
003\_cracker\_box        & 46.7 & \textbf{49.9} & 44.4    & 49.3          & 48.9             & 47.5   & 48.9          \\
004\_sugar\_box          & 37.7 & 36.3          & 35.5    & \textbf{43.6} & 43.1             & 39.3   & 40.9          \\
005\_tomato\_soup\_can   & 30.2 & 30.4          & 28.5    & \textbf{36.8} & 32.4             & 31.5   & 33.1          \\
006\_mustard\_bottle     & 40.4 & 42.6          & 38.1    & \textbf{45.7} & 44.4             & 44.6   & 43.2          \\
007\_tuna\_fish\_can     & 19.8 & 17.9          & 15.1    & 20.4          & \textbf{21.7}    & 20.0   & 19.5          \\
008\_pudding\_box        & 25.8 & 24.9          & 21.7    & \textbf{31.7} & 31.6             & 29.4   & 30.6          \\
009\_gelatin\_box        & 20.4 & 20.3          & 16.2    & 23.5          & \textbf{23.5}    & 22.2   & 20.4          \\
010\_potted\_meat\_can   & 28.2 & 27.6          & 25.1    & \textbf{34.7} & 33.2             & 33.2   & 31.9          \\
011\_banana              & 16.6 & 9.9           & 10.5    & 13.9          & 21.4             & 22.0   & \textbf{23.0} \\
019\_pitcher\_base       & 49.1 & 48.0          & 43.5    & 48.1          & 50.2             & 49.2   & \textbf{51.4} \\
021\_bleach\_cleanser    & 45.1 & 43.1          & 41.1    & \textbf{48.4} & 47.9             & 44.3   & 48.0          \\
\it{024\_bowl}                & 61.5 & 64.5          & 62.1    & \textbf{71.8} & 66.6             & 64.6   & 66.4          \\
\textit{025\_mug}                 & 32.5 & 31.5          & 26.2    & 34.9          & 35.4             & 34.0   & \textbf{35.8} \\
035\_power\_drill        & 37.5 & 19.7          & 31.6    & 37.0          & 38.5             & 35.6   & \textbf{39.5} \\
\it{036\_wood\_block}         & 70.1 & \textbf{76.2} & 67.4    & 75.1          & 73.4             & 72.2   & 71.8          \\
037\_scissors            & 12.5 & 8.6           & 10.2    & 12.5          & 15.4             & 14.6   & \textbf{15.6} \\
\textit{040\_large\_marker}       & 8.6  & 4.5           & 4.3     & 5.9           & 9.2              & 10.2   & \textbf{11.8} \\
\it{051\_large\_clamp}        & 44.5 & 42.4          & 43.8    & \textbf{53.5} & 50.3             & 51.3   & 51.5          \\
\it{052\_extra\_large\_clamp} & 49.2 & 39.8          & 50.3    & 52.0          & \textbf{55.7}    & 53.8   & 53.1          \\
\it{061\_foam\_brick}         & 37.2 & 43.1          & 33.1    & \textbf{50.5} & 43.3             & 42.8   & 43.0          \\ \midrule
                         & 36.3 & 34.7          & 31.8    & 39.0          & \textbf{40.0}    & 38.7   & 39.5          \\ \bottomrule
\end{tabular}
\caption{We report the performance of our proposed methods on a per-object basis, comparing the accuracy of a single multi-object model for both mono and stereo configurations using the common ADD0.1 metric. Our findings show that the integration of stereo vision significantly enhances performance on both SO-Pose and GDRN(PP).
Stereo-GDRN with early fusion increases overall performance over mono GDRN from 36.3 to 38.7. Adding disparity prediction using the shared backbone adds another 1.3. Double fusion, entailing fusion both early and late in the architecture, places itself between both early fusion variants. While it demonstrated superior performance for specific objects, it did not consistently enhance overall accuracy over early fusion only. Best result per object in \textbf{bold font}, symmetric objects in \it{italics}.}
\label{tab:results}
\end{table*}

\textbf{Optimization.}
To optimize our model, we employed the Ranger \cite{lessw2019ranger} optimizer with a learning rate of $0.0001$. The learning rate undergoes a warm-up phase with a factor of 0.001 during the initial 1000 iterations. In the final 27\% to 72\% of the training period we use learning rate annealing using the cosine function. 

\subsection{Results}

We evaluate our methods on the Stereo PBR YCB-V DS dataset, as its perfect labeling presents itself for fair comparison between the algorithms. Table \ref{tab:results} compares our approaches SO-Stereo and GDRN-Stereo to GDRN, GDRNPP, and SO-Pose. In our experiments we find that some of our ideas such as mid or late fusion do not lead to a meaningful increase in performance when accounting for the additional compute requirements. Therefore we only report results on a selection of our approaches. 

Our findings are that GDRN-Stereo with early and double fusion leads to best results, with the combination of early fusion with disparity features outperforming other methods. SO-Pose performed best with early fusion, but does not match the best performing versions of GDRN-Stereo.

Interestingly the best approach, GDRN-Stereo with incorporating early fusion with disparity, does not outperform by performing best on a large number of objects, but by being less bad on the worst performing objects (comparing GDRN with GDRN-Stereo and objects 040\_large\_marker and 037\_scissors). In contrast, SO-Stereo with early fusion achieves best per-object performance for 9 out of 21 objects, but not best performance across all objects. We also note that, while we see a big performance difference between the mono versions of GDRN and SO-Pose, the gap narrows with our stereo extension. Especially impressive is the performance gain when integrating stereo into SO-Pose. Also noteworthy is the strong outperformance on symmetric objects using our methods, especially with early fusion SO-Stereo. The single worst object throughout all algorithms is the, contrary to its name, small 040\_large\_marker, with single decimal results for all monocular methods, but we can report noticeably improvements using our GDRN-Stereo. Among the best performing objects are 036\_wood\_block and 024\_bowl, with the first being one of only two objects on which a single-view algorithm outperformed our stereo extensions. 

In sum, all of our stereo methods outperform the monocular counterparts in the single model multi-object scenario on our heavily occluded YCB-V DS dataset with a significant margin.

\section{Conclusion}
We extended single-view state-of-the-art approaches for 6D pose estimation, GDRNet(PP) and SO-Pose, to incorporate stereo information and doing so, improved upon the performance noticeably. Our approach can be integrated in similar algorithms and adds only limited compute overhead. We provide our stereo + depth \href{https://huggingface.co/datasets/tpoellabauer/YCB-V-DS}{YCB-V dataset (YCB-V DS)} to be used in future work. 

\bibliographystyle{splncs04}
\bibliography{bib}

\end{document}